# Dragon-Alpha&cu32：
# A Java-based Tensor Computing Framework With its High-Performance CUDA Library

Zhiyi Zhang, Pengfei Zhang, Qi Wang

***Abstract*—Java is very powerful, but in Deep Learning field, its capabilities probably hasn't been sufficiently exploited. Compared to the Java-based deep-learning-frameworks, the Python-based (PyTorch, TensorFlow, etc) are undoubtedly the mainstream, due to their easy-to-use, flexibility and better ecosystem. Dragon-Alpha is a Java-based Tensor Computing Framework, with easy-to-use, high-scalability and high-performance, trying to break Java's dilemma in deep learning field and make it more effective. Dragon-Alpha supports different levels of APIs, and can be used as a deep-learning-framework through its user-friendly high-level APIs. Dragon-Alpha has potential to aggregate computing-power across heterogeneous platforms and devices, based on its multi-layer architecture and Java's big-data ecosystem. Dragon-Alpha has its asynchronized APIs to improve parallelism, and highly-optimized CUDA library 'cu32' which adopts unique convolution\deconvolution operators for 'small feature maps'. The experiments show that, compared to 'PyTorch&cuDNN', 'Dragon-Alpha&cu32' costs less time and memory ($75.38\% \rightarrow 97.32\%$, $29.2\% \rightarrow 66.4\%$), to train some typical neural networks (AlexNet, VGG, GoogleNet, ResNet) on Cifar-10.**

## I. INTRODUCTION

NOWADAYS, Deep Learning[1] (DL) is obviously the hottest field of Artificial Intelligence. There are many DL frameworks, like TensorFlow[19], PyTorch[20], Caffe[18], and DeepLearning4j[21]. All the above mentioned frameworks are based on Python&C++, except for DeepLearning4j based on Java. Compared to DeepLearnig4j, PyTorch supports dynamic graphs, costs less code to execute neural networks with concise APIs, and earns much more users. Broadly, Python is amazingly hot in DL, while Java is relatively much unpopular. To list some reasons, the Python-based DL frameworks are easier to learn and use, and more flexible than the Java-based in some cases. Additionally, Python has many related libraries to build an outstanding DL ecosystem.

But Java has its great potential in DL. As the most widely-used programming language, Java has lots of users, projects and communities. Java has good performance, much better than Python and sometimes close to C++. Java is first-class in flexibility, based on its object-oriented characteristic and reflecting-mechanism. With the best big-data ecosystem, Java is able to aggregate massive computing-power to meet the need for training large models. Contrast to Python, Java's complexity is shortage, but also an advantage to provide more selections. Since the traditional Java programming style is a little cumbersome for DL, it's required to adopt some innovative designs to enhance Java's usability. Conclusively, it's meaningful to build an open-source Java-based DL framework, that is high-performance, highly-scalable and as-easy-to-use-as-Python.

In DL field, GPU is one of the most important types of hardware, especially for high-complexity operations, such as convolution and matrix-multiplication. NVIDIA cuDNN[17] is the most widely-used GPU library (lib) for DL, and also an indispensable element for many DL frameworks to reach high-performance. But cuDNN[17] is not open-source, we still need an open-source one, to make some contributions and guidance for GPU programming and parallel computation.

To address such problems, Dragon-Alpha&cu32 is presented. Dragon-Alpha, abbreviated as Alpha, is a Java-based Tensor Computing Framework, and can be used to express and execute DL algorithms. As an efficient GPU library, cu32 is integrated to Alpha at the bottom. To run Alpha's applications, you only need an appropriate version of JDK and CUDA[2][3].

The code of Alpha&cu32, and the experimental-data can be download at https://github.com/GilgameshXYZ123/Dragon-Alpha.

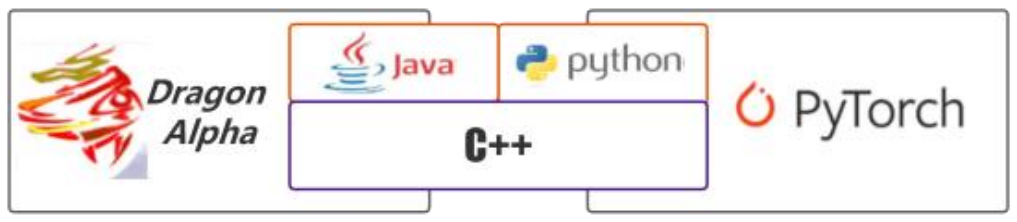

**Fig. 1.** Alpha and PyTorch are both implemented by C++ at the bottom. Alpha is implemented by Java but PyTorch is implemented by Python at the top.

This research was supported by National Natural Science Foundation of China (No. 32070399); Anhui Science and Technology Major Project (No. 202103a06020014), the Chinese Academy of Sciences-Henan Province Achievement transfer and Transformation Project (No. 2022208); Hefei Science and Technology Project (No. 2021GJ065).

Zhiyi Zhang, Pengfei Zhang and Qi Wang are with the HFIFPS (Hefei Institutes of Physical Science, Chinese Academy of Sciences), Hefei, China. Zhiyi Zhang is also with USTC (University of Science and Technology of China), Hefei, China.

Zhiyi Zhang (email: gilgamesh@mail.ustc.edu.cn) is the first author, Pengfei Zhang (email: pfzhang@aiofm.ac.cn) is the second author, and Qi Wang (email: wangqi@ipp.ac.cn) is the corresponding author.

```java
public static class Block extends Module {
    Unit conv1, bn1, downsample;
    public Block(int in_channel, int out_channel, int stride) {
        conv1 = nn.conv3D(false, in_channel, out_channel, 3, stride, 1);
        bn1 = nn.batchNorm(false, out_channel);
        if(stride != 1 || out_channel != in_channel)
            downsample = nn.sequence(
                    nn.conv3D(false, in_channel, out_channel, 3, stride, 1),
                    nn.batchNorm(out_channel));
    }

    @Override
    public Tensor[] __forward__(Tensor... X) {
        Tensor[] res = X;
        X = F.leakyRelu(bn1.forward(conv1.forward(X)));
        if(downsample != null) res = downsample.forward(res);
        X = F.add(X[0], res[0]);
        return F.leakyRelu(X);
    }
}
```

```java
public static class Net extends Module {
    Unit conv1 = nn.conv3D(false, 3, 64, 3, 1, 1);
    Unit bn1 = nn.batchNorm(false, 64);
    Block block1 = new Block(64, 128, 2);
    Block block2 = new Block(128, 256, 2);
    Unit fc = nn.fullconnect(true, 256, 10);

    @Override
    public Tensor[] __forward__(Tensor... X)  {
        X = bn1.forward(conv1.forward(X));
        X = F.leakyRelu(X);

        X = block1.forward(X);
        X = block2.forward(X);

        X = F.adaptive_maxPool2D(1, X);
        X = fc.forward(F.flaten(X));
        return X;
    }
}
```

```java
Net net = new Net().train().init(eg).println();
Optimizer opt = alpha.optim.Adam(net.params(), lr);
LossFunction loss = alpha.loss.softmax_crossEntropy();
BufferedTensorIter iter = Cifar10.train().buffered_iter(eg, batch_size);

eg.sync(false).check(false);
for(int i=0; i<epoch; i++)
    for(iter.reset(true); iter.hasNext();) {
        Pair<Tensor, Tensor> pair = iter.next();
        Tensor x = pair.input;
        Tensor y = pair.label;

        Tensor yh = net.forward(x)[0];
        System.out.println("loss = " + loss.loss(yh, y));
        net.backward(loss.gradient(yh, y));
        opt.update().clear_grads();
        net.gc();
    }
```

**Fig. 2.** Alpha's high-level APIs

## II. CHARACTERISTICS

The most significant characteristics of Alpha are as follows.

**Different levels of APIs** Alpha supports not only user-friendly high-level APIs, but also more complex low-level APIs to avoid the cost of automatic-differentiation, tensor-encapsulation, memory-pooling, params-check, etc. From the perspective of the highest-level APIs, Alpha can be regarded as a DL framework. Besides, through the lower-levels APIs, it's feasible to make other kinds of applications, even a different DL framework.

**Easy-to-use** At Alpha's top levels, the format of operators are close to math expressions, and the APIs are neat and have some similarities to PyTorch's, although their underlying implementations are different. Alpha's coding-style is between Java's and Python's, to make it familiar for both Java and Python users. The 'method chaining' and 'factory pattern' are widely adopted for convenience. Instead of *getter*&*setter*, Alpha uses a more concise function-naming-rule to facilitate property management. Some dangerous operations will be allowed if the related properties are configured, such as closing params-check, turning on async-mode, operating memory-addresses, modifying computing strategies, etc.

To build up a dynamic acyclic computational graph, it only needs to implement a subclass of *Module* by overwriting the *constructor* and *forward* method. After forward propagation, use the *backward* method with the loss function to find gradients, then use the optimizer for gradient descent\ascent[7]. Exceptions will be thrown if the computational graph is changed during backward propagation. A related example is shown in Fig. 2.

**High-scalability** Alpha is designed to cross heterogeneous platforms and devices. As a Java application, Alpha is independent of platforms because of JVM, and can be integrated into Java's big-data frameworks (Hadoop, Flume, Spark, etc) naturally to construct distributed applications. Benefiting from the muti-layer architecture, Alpha's bottom layers can shield the concrete details of devices from the higher layers. For different kinds of devices, different kinds of bottom layers can be implemented to achieve polymorphism. Alpha can use specific devices through the corresponding bottom layers, while the higher layers stay unchanged.

**Asynchronized APIs** Alpha's asynchronized (async) APIs enable programmers to execute a set of operators in parallel on devices, and use callback functions as semaphores to wait for the operators' end and attach some actions in need. Programmers only need to focus on the correlation among operators, regardless of the underlying details to implement parallelism. In addition, CPU is able to do something instead of just waiting, while devices are busy working. On CUDA platforms, one CPU thread can launch multiple kernels in different streams without being blocked. A relevant example is shown in Fig. 3.

**High-performance** In Alpha, the matrix-multiply and convolution\deconvolution (conv\deconv) operators are highly optimized. The last dimension of tensors is transparently padded to $4x$, to increase the memory bandwidth, by using 128bit as the minimum unit of memory-access.

Based on the async APIs, the utilization of devices can be improved by executing multi operators in parallel along with some CPU instructions (e.g., the data-prepossessing and model-training can be performed concurrently).

To avoid the overhead of frequent memory-allocation through system calls, the memory-pool is built for holding and reusing memory-blocks, based on the abstract malloc&free methods provided by the lower-level APIs. A certain amount of pinned memory is managed to perform as a cache between CPU and devices, to accelerate data-transmitting by direct-memory-access.

To reduce the volume of data-transmission, pictures are stored and transferred in *int8* data-type, and convert to a float data-type (usually *float32*) at the destination, through an operator with $O(n)$ complexity.

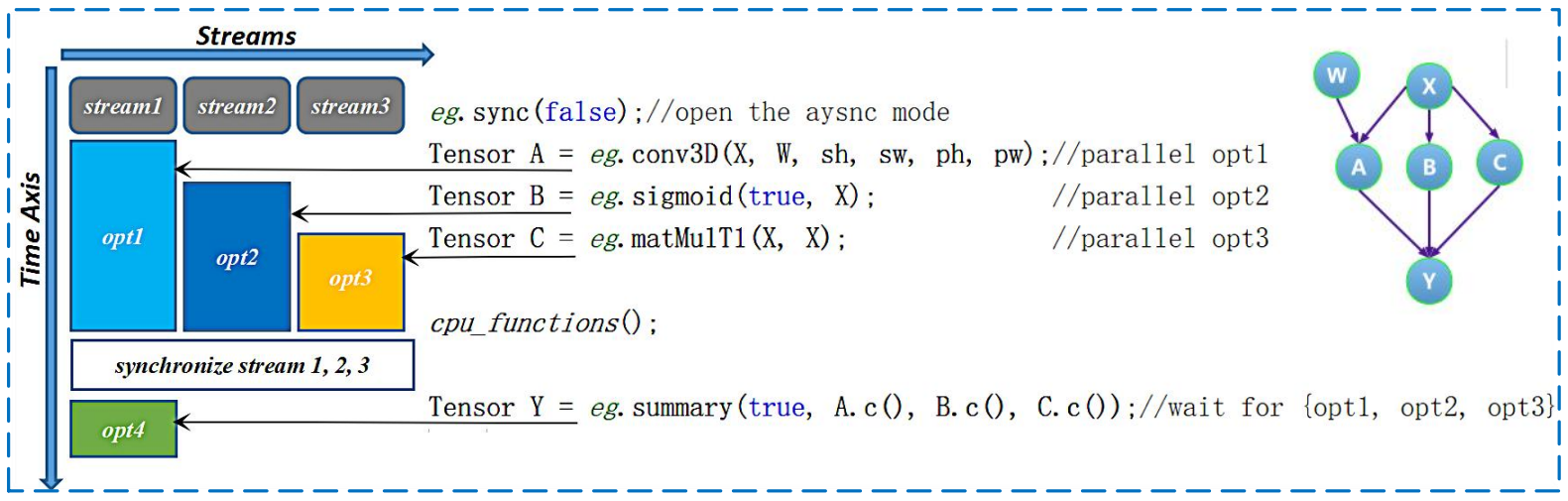

**Fig. 3.** Alpha's Async APIs
$\{opt1_{conv3D}, opt2_{sigmoid}, opt3_{matMulT1}\}$ are submitted to $\{stream1, stream2, stream3\}$ one by one, and are executed in parallel along with some CPU functions. Before submitting $opt4_{summary}$ to $stream1$, use $Tensor.c$ method to synchronize $\{stream1, stream2, stream3\}$, to ensure the completion of $\{opt1, opt2, opt3\}$

## III. ARCHITECTURE

The Dragon-Alpha consists 6 layers, as shown in Fig 4.

**Layer-1 Native libs** The native libs contains the most underlying and sophisticated computing logics and strategies. They are implemented by C++ to get high-performance.

For Alpha, 'cu32' is a connected GPU native lib, aimed at single-precision float (*float32*) operations on CUDA. cu32 consists of 13 dlls, and takes the most effort to optimize the conv\deconv operators, especially for 'big channel and small feature maps'. cu32 has been tested on RTX 3060ti and GTX 1050. When the GPU is running at almost full capacity, under the promise of accuracy, cu32 has similar performance to cuDNN[17], for many common situations of DL.

**Layer-2 EngineBase** EngineBase is an abstract class, which defines a set of abstract primitives by abstract methods. Based on polymorphism, different sub-classes of EngineBase can be implemented for specific devices, such as CPU, GPU, TPU, FGPA, etc. The CudaFloat32EngineBase is a subclass of EngineBase for *float32* operations on CUDA, and solves logics and strategies in higher levels compared to cu32. Through JNI, the Java methods of CudaFloat32EngineBase are mapped to the C++ functions in cu32.

**Layer-3 EngineCore** EngineCore is a higher-level encapsulation of EngineBase, works based on EngineBase's primitives, and additionally has functions like memory-pooling, params-check and error-handling. The details of device are transparent to EngineCore, since they are shielded by EngineBase. EngineCore can work in a specific situation, if a suitable EngineBase is configured. For example, EngineCore can directly use primitives of the configured CudaFloat32EngineBase, to execute operators on GPU.

**Layer-4 Engine** Engine is a higher-level encapsulation of EngineCore, through EngineCore's primitives. EngineCore's operators are inconvenient to use, because the input params are numerous and datas are represented as 64bit-addresses. Engine's operators are much easier to use, the in\outputs and data-addresses are encapsulated to Tensors, which have some useful mechanisms like memory-padding, saving\transmitting information, managing gradients, acting as semaphores for async APIs. The *delete* method is used to release the bound memory-blocks of Tensors, and will also be called when Tensors are garbage-collected by JVM.

'sync' and 'check' are two important properties of Engine. If set 'sync' to false, the async-mode will be enabled: operators immediately return the resultant Tensors, regardless of whether they are ended. To ensure the operations are completed, *Tensor.c* method can be used to block related threads. If set 'check' to false, params-check will be banned. Among the loops of training neural networks, the computational graph and params often remain static or changes within a fixed range. As a result, to ensure correctness, it's sufficient to check params at the first few times, but wasteful to check the duplicate params every time. After certain iterations to ensure the correctness and stability, it's recommended to open the async-mode and close params-check to improve efficiency.

**Layer-5 Unit, etc** Unit is a higher-level encapsulation of Engine, and performs as a node in computational graph. A complex Unit has its inner Units to form a sub graph. A Unit must be initialized with an Engine and an optional hierarchical name. The core methods of Unit are *forward* and *backward*, whose in\outputs are both Tensor arrays. The *forward* corresponds to forward propagation, and the *backward* finds gradients based on chain-rule with the feedback of loss function. The automatic-differentiation is implemented like routers. In forward propagation, Tensors carry information to the subsequent Units and receive their feedback, to solve the directed edges of the graph. In backward propagation[6], based on the solved edges, Units collect and sum up gradients from the subsequent, then find gradients for their precursors. As the garbage-collection mechanism of JVM is heavy and usually lags behind, it's advised to use *Unit.gc* method with smaller cost, to recycle unnecessary memory-blocks timelier after each mini-batch is processed.

LossFunction, Optimizer, DataSet, Stat and etc, are offered to complete the training process of neural networks. Some methods for basic image\file processing are provided by DragonCV\DragonFL.

**Layer-6 Alpha** *alpha* packaged almost everything in Dragon-Alpha. By using keywords like *alpha.nn*, *alpha.F*, *alpha.optim*, *alpha.loss*, *alpha.engine*, etc, it's convenient to build various kinds of directed acyclic computational graphs.

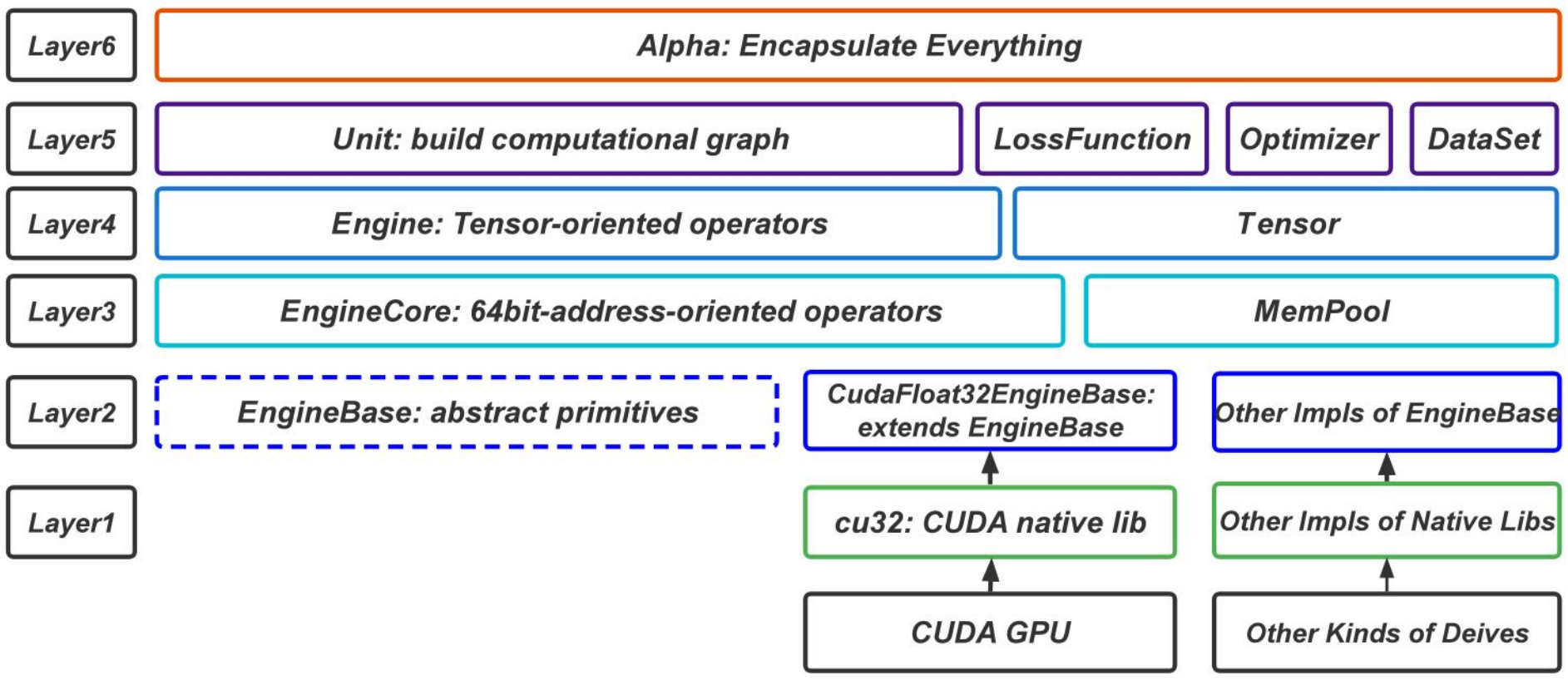

**Fig. 4.** Alpha's multi-layer architecture

## IV. EXPERIMENT

To evaluate Alpha's performance, PyTorch[20] (v1.7.1) is used as a benchmark, and can also be used as a bridge to indirectly compare Alpha with other DL frameworks. Some typical kinds of neural networks were created, executed and evaluated on Cifar-10, using both Alpha and PyTorch.

The variables were controlled, to ensure the networks created by Alpha or PyTorch are logically the same, and were initialized, trained, tested under the same conditions 5 times. The experimental conditions are listed in TABLE I. The metrics listed in TABLE II were tracked, recorded and averaged to make a comparison. The performance of Alpha and PyTorch are listed in TABLE III, and the loss function curves are shown in Fig.5-12.

The experimental results verify the correctness and efficiency of Alpha. Except for 'ResNet&SGD', Alpha and PyTorch have similar loss function curves and accuracy. On all networks, Alpha has advantage over PyTorch in speed and memory-cost, and spends 75.38% → 97.32% time and 29.2% → 66.4% GPU-memory of PyTorch's.

Alpha has higher speed on Cifar-10, and the reasons could be as follows. cu32 is sufficiently optimized based on CUDA's characteristics, and has unique deconv\conv operators to reduce the computational complexity for 'small feature maps'. Typically, due to down-sampling, with going deeper, the channels increase, but the feature maps become smaller and smaller. Such unique operators will be selected, if feature maps are smaller than a certain threshold. The smaller the feature maps, the more complexity such operators reduce. Through the async APIs, Alpha is able to execute multi operators concurrently in some work stages, which improves GPU-utilization and hid the execution of some CPU functions, while PyTorch usually uses one workflow.

PyTorch's GPU-utilization is higher, mainly because cuDNN can schedule GPU resources more fully than cu32, rather than due to high-level software designs. For ResNet, when the input-feature-size $\geq 128 \times 128$, cuDNN[17] is faster than cu32, and the GPU-utilization of cu32 ($90\% \rightarrow 99\%$) shakes more violently than cuDNN ($96\% \rightarrow 99\%$). In such case, cu32's operators perform less effectively to reduce computational complexity, due to the increase of feature maps. cu32 needs some updates for large feature maps.

Alpha has more in-place operators than PyTorch (in-place BatchNorm, LayerNorm, etc), and tries to calculate gradients directly on the memory-space of the precursors. In both forward and backward propagation, Tensors no longer need will be released implicitly by Units through *Tensor.c* method. Such mechanisms allow Alpha to reuse memory efficiently, and could be reasons for Alpha's lower memory-cost.

Since it's relatively complex to solve BatchNorm's gradients, the related implementations of cu32 and cuDNN could be quite different, which may cause different accuracy due to the limited precision of float data-types. Both ResNet18[12] and ResNet34[12] use many BatchNorm[5] layers, which may magnify the difference of gradients between Alpha and PyTorch, to cause bigger discrepancies in parameter-hyperplane and loss function curves than other networks'. Such difference will decrease, if remove certain BatchNorm layers in ResNet.

TABLE I
Experimental Conditions

| | |
|---|---|
| ***Platform*** | RTX 3060ti GPU, CUDA v11.3. |
| ***Execution*** | All computations uses float32 data-type. At the same time, only run one network created by PyTorch or Alpha, and ensure the network can grasp almost all GPU resources, without interference from other programs. |
| ***Input feature maps*** | $32 \times 32 \times 3$ pixel input-feature-maps. Collect N (batchSize =512) feature-maps and divide all pixels by 255.0, to obtain a corresponding 4D tensor in float32 data-type. The main order of the 4D tensors for Alpha is $[N, H, W, C]$, and for PyTorch is $[N, C, H, W]$. |
| ***Labels*** | 10-class labels, using one-hot encoding method. |
| ***Loss Function*** | Softmax with CrossEntropy. |
| ***Optimizer*** | $lr = 0.01$. Adam[15] ($\beta_1 = 0.9, \beta_2 = 0.999, \varepsilon = 10^{-8}$) or SGD[13]. |
| ***BatchNorm[5]*** | The hyper-parameter is ($\varepsilon = 10^{-8}, \beta = 0.1, affine = true$) . For AlexNet[9], VGG[10] and GoogleNet[11], BatchNorm layers are introduced to accelerate convergence and solve the problem of gradients-vanishing. |
| ***Initialization*** | Use Kaiming-uniform[22] to initialize the convolution and full-connect layers. For BatchNorm[5] layers, their weights and bias are initialized to 1 and 0 respectively. |

TABLE II
Metrics to be tracked

| | |
|---|---|
| ***Loss function curve*** | The value of loss function is recorded per 10 iterations, to generate a time sequence. For a specific network, collect 5 such sequences, and draw the loss function curve based on the element-wise average of the 5. |
| ***Total training time*** | Exclude the time of initializing and saving the model. The timing starts from the beginning of the first forward propagation, to the end of the last optimizing. |
| ***Accuracy*** | On both train\test set. Take the average of the 5 experiments for each network. |
| ***GPU memory-cost & utilization*** | Through 'nvidia-smi', record the maximum memory-cost and utilization of GPU. Since the memory-cost will converge to its maximum after several iterations. Also, the fluctuation of utilization is very small, resulting in a tiny difference of only a few percentage points between the minimum and maximum. |

TABLE III
Performance of Alpha and PyTorch

In each cell, the first line is for Alpha, the the second is for PyTorch

| Network& Optimizer | Total-training-time / epoch | Alpha / PyTorch | GPU memory-cost | Alpha / PyTorch | GPU utilization | Accuracy on train\test set |
|---|---|---|---|---|---|---|
| **AlexNet&Adam** | $\frac{72.638\ s}{20\ epoch} = 3.6319\ s/epoch$<br>$\frac{74.639\ s}{20\ epoch} = 3.7319\ s/epoch$ | 97.32% | $885_{MB}$<br>$3021_{MB}$ | 29.20% | 95%<br>97% | 96.20% 73.97%<br>95.85% 74.17% |
| **VGG16&Adam** | $\frac{206.554\ s}{20\ epoch} = 10.328\ s/epoch$<br>$\frac{220.322\ s}{20\ epoch} = 11.016\ s/epoch$ | 93.75% | $2193_{MB}$<br>$3872_{MB}$ | 56.64% | 97%<br>99% | 97.97% 82.01%<br>97.64% 81.75% |
| **VGG19&Adam** | $\frac{310.439\ s}{25\ epoch} = 12.418\ s/epoch$<br>$\frac{325.622\ s}{25\ epoch} = 13.025\ s/epoch$ | 95.34% | $2348_{MB}$<br>$3984_{MB}$ | 58.93% | 97%<br>99% | 97.93% 80.78%<br>97.32% 81.23% |
| **GoogleNet&Adam** | $\frac{236.765\ s}{25\ epoch} = 9.471\ s/epoch$<br>$\frac{314.073\ s}{25\ epoch} = 12.563\ s/epoch$ | 75.38% | $2508_{MB}$<br>$4765_{MB}$ | 52.63% | 90%<br>98% | 95.63% 76.51%<br>93.83% 76.67% |
| **ResNet18&Adam** | $\frac{161.470\ s}{25\ epoch} = 6.4588\ s/epoch$<br>$\frac{181.063\ s}{25\ epoch} = 7.2425\ s/epoch$ | 89.18% | $1517_{MB}$<br>$2750_{MB}$ | 55.16% | 91%<br>98% | 98.99% 75.10%<br>99.31% 78.33% |
| **ResNet18&SGD** | $\frac{315.905\ s}{50\ epoch} = 6.3181\ s/epoch$<br>$\frac{346.038\ s}{50\ epoch} = 6.9207\ s/epoch$ | 91.29% | $1331_{MB}$<br>$2699_{MB}$ | 49.31% | 91%<br>98% | 99.77% 42.57%<br>83.26% 53.29% |
| **ResNet34&Adam** | $\frac{317.385\ s}{25\ epoch} = 12.695\ s/epoch$<br>$\frac{349.692\ s}{25\ epoch} = 13.988\ s/epoch$ | 90.76% | $2263_{MB}$<br>$3408_{MB}$ | 66.40% | 93%<br>98% | 97.66% 76.77%<br>98.39% 79.73% |
| **ResNet34&SGD** | $\frac{629.303\ s}{50\ epoch} = 12.586\ s/epoch$<br>$\frac{672.015\ s}{50\ epoch} = 13.440\ s/epoch$ | 93.65% | $1981_{MB}$<br>$3291_{MB}$ | 57.46% | 92%<br>98% | 88.03% 42.42%<br>99.99% 43.99% |

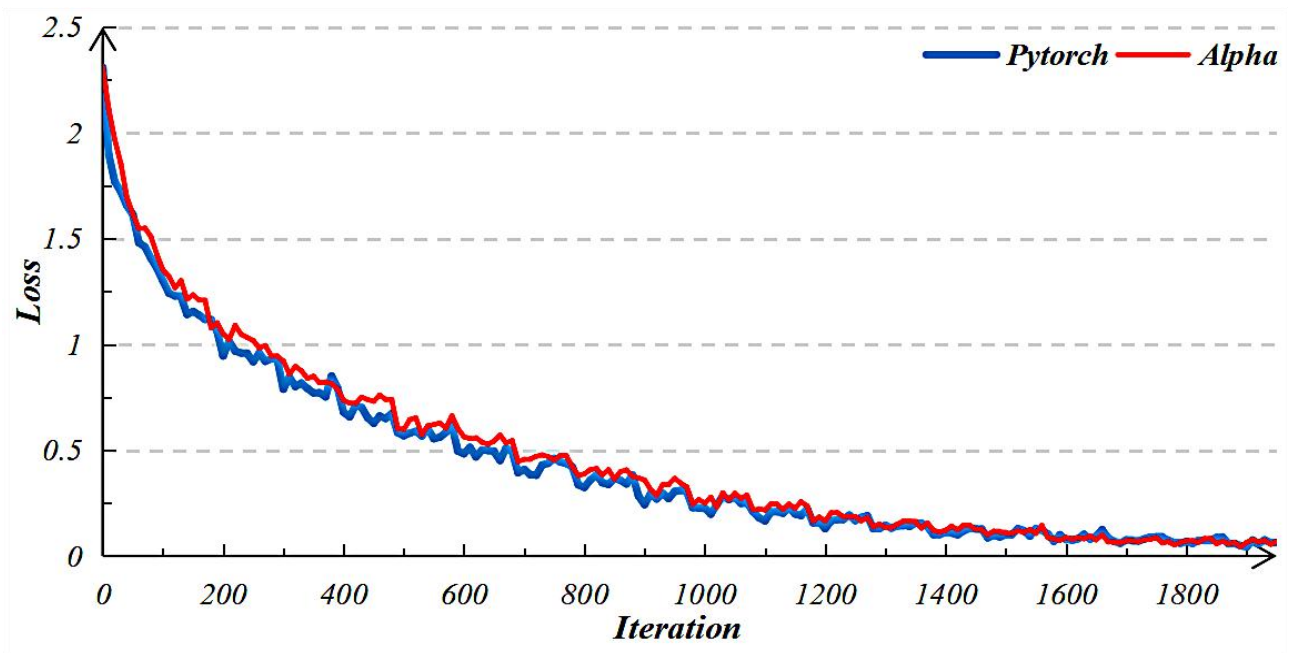

**Fig. 5.** loss function curve: AlexNet&Adam for 20 epochs

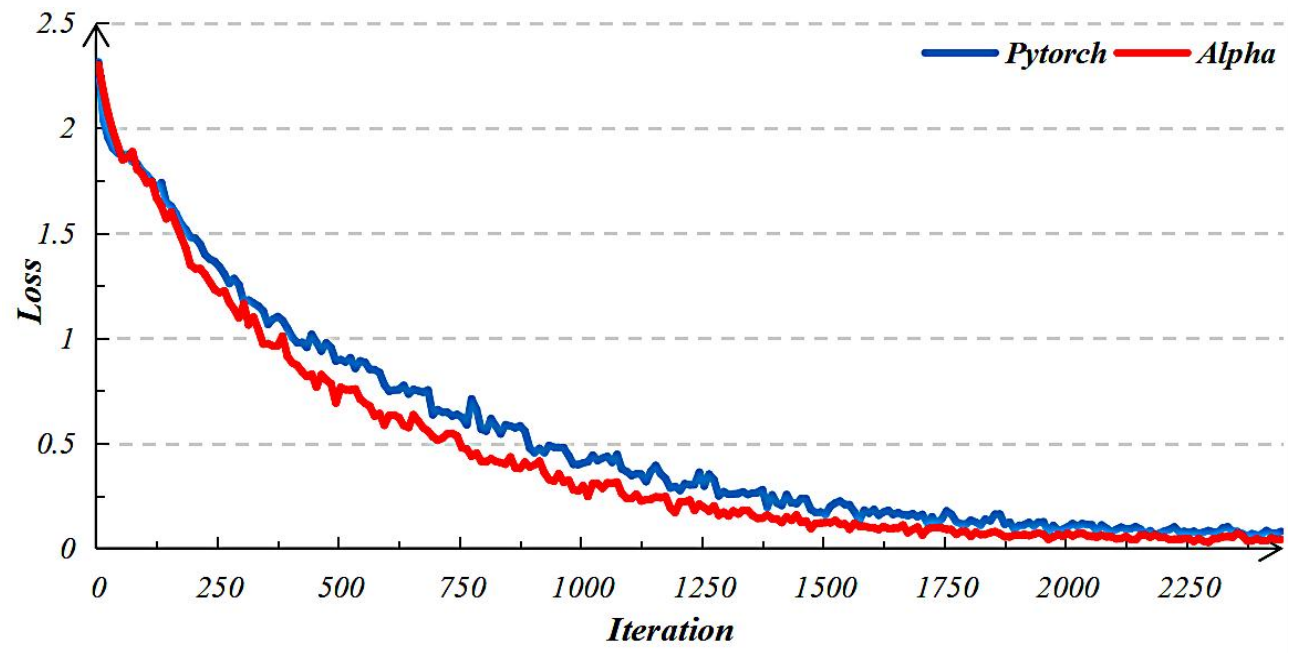

**Fig. 7.** loss function curve: VGG19&Adam for 25 epochs

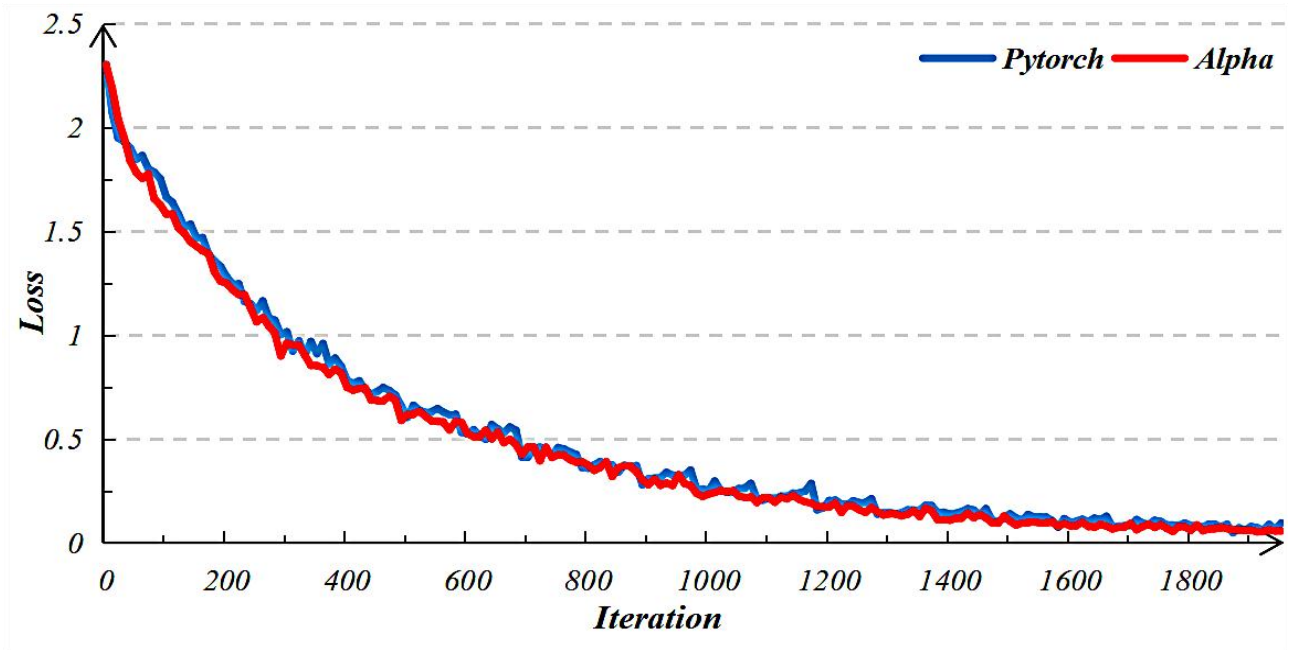

**Fig. 6.** loss function curve: VGG16&Adam for 20 epoch

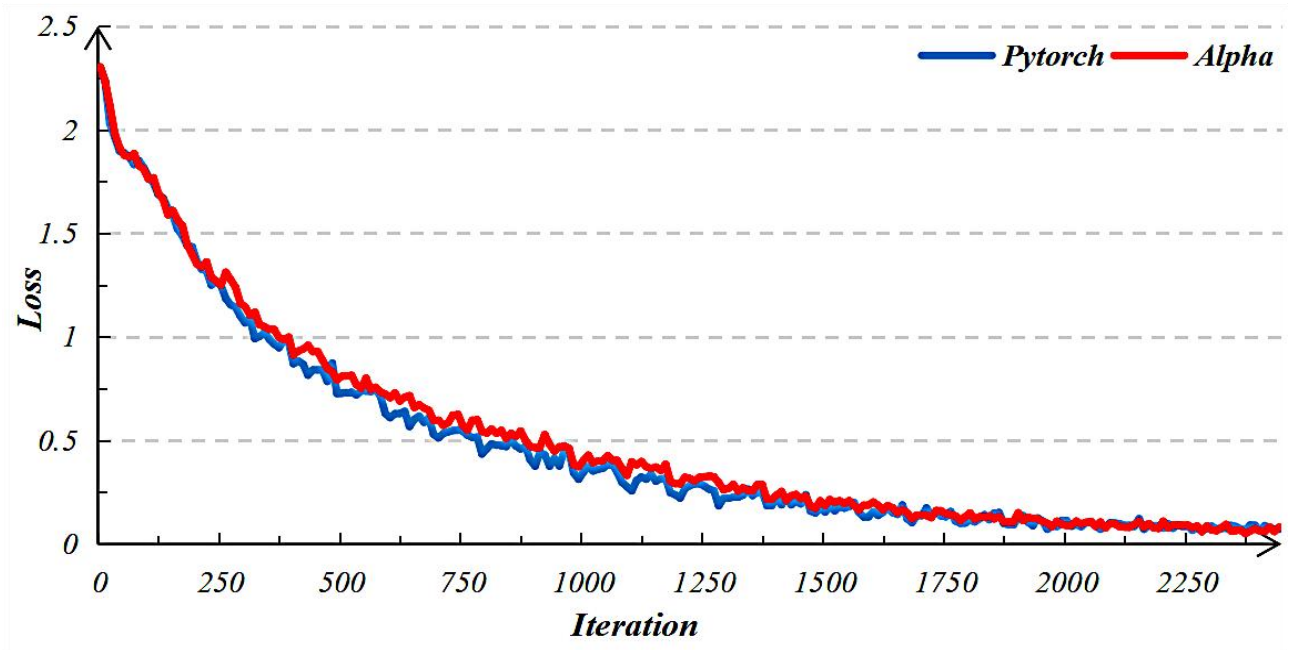

**Fig. 8.** loss function curve: GoogleNet&Adam for 25 epochs

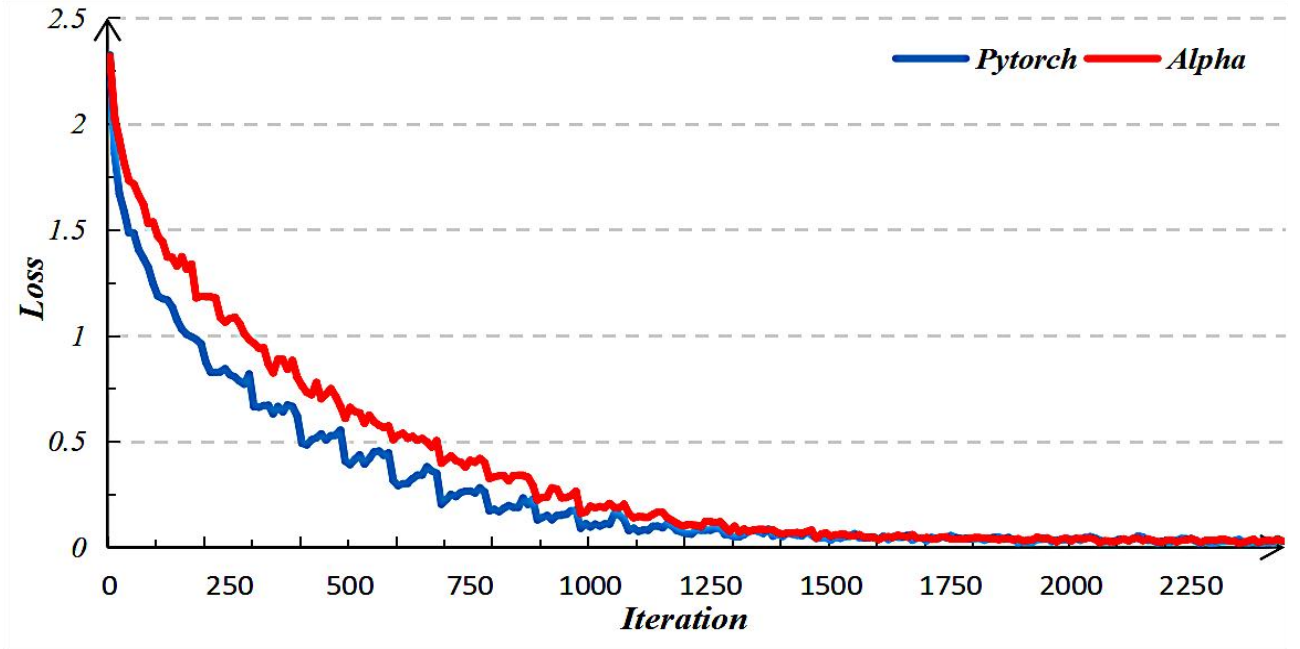

**Fig. 9.** loss function curve: RestNet18&Adam for 25 epochs

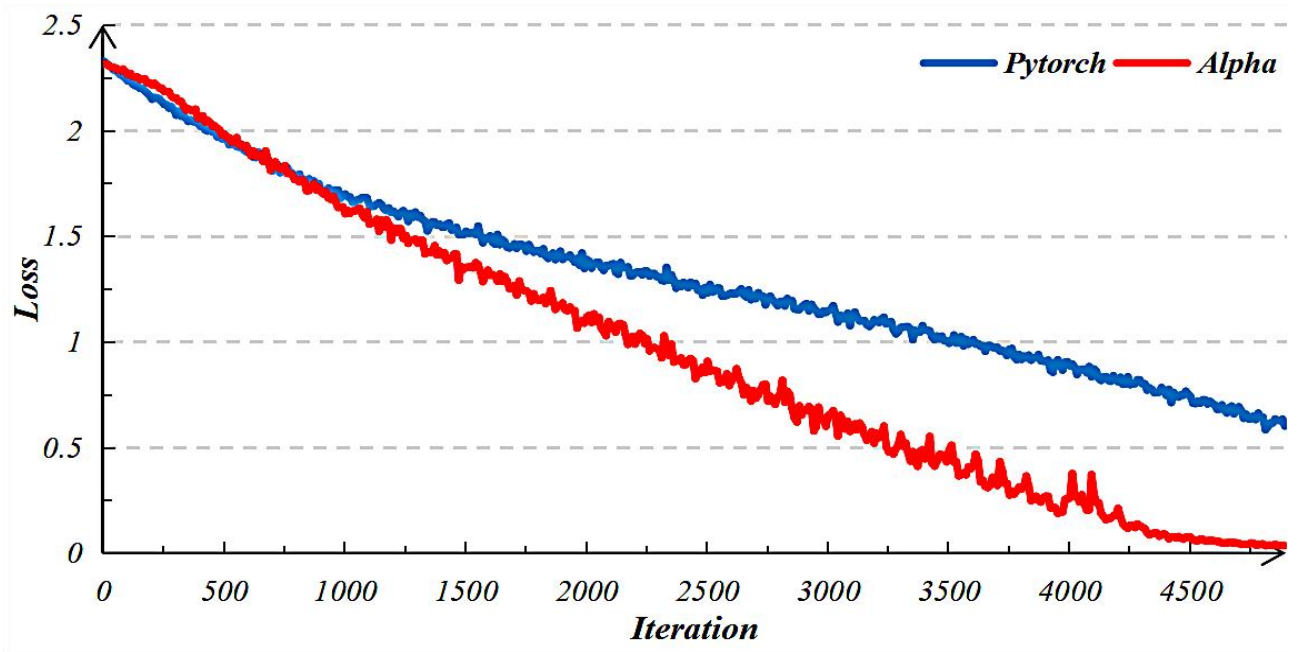

**Fig. 10.** loss function curve: ResNet18&SGD for 50 epochs

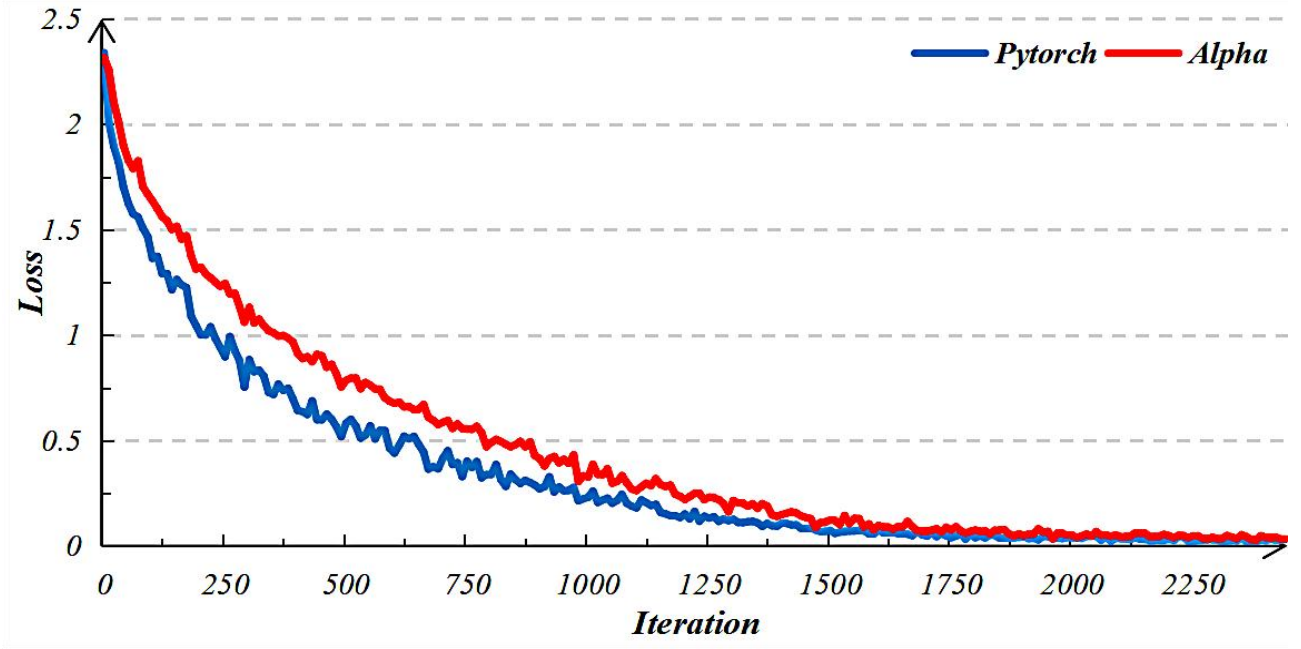

**Fig. 11.** loss function curve: ResNet34&Adam for 25 epochs

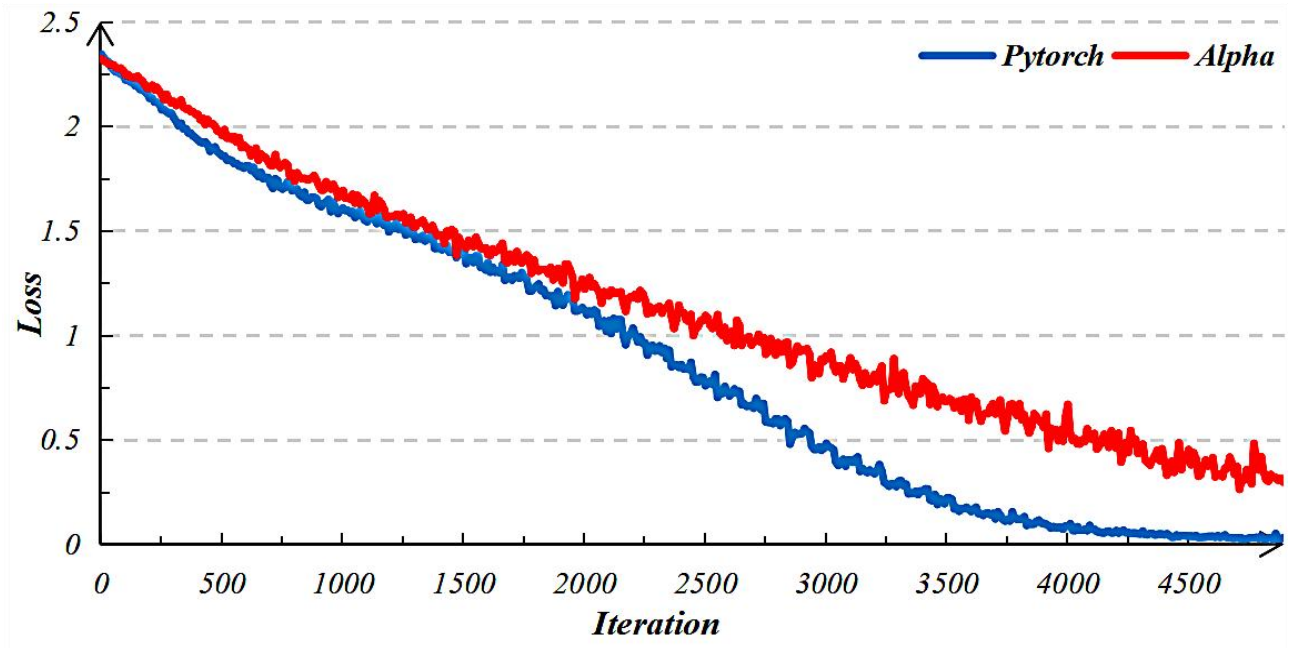

**Fig. 12.** loss function curve: ResNet34&SGD for 50 epoch

## V. Conclusion

This paper discusses the background, characteristics, architecture and experiments of Dragon-Alpha. Alpha tries to raise Java's effectiveness in DL field, by taking its advantages. Alpha adopts some innovative designs to become easy-to-use, and realize high-performance through the async APIs and cu32. Based-on its multi-layer architecture and Java big-data frameworks, Alpha is highly-scalable and capable of aggregating heterogeneous computing-resources. By using PyTorch as a benchmark in experiments, the correctness and efficiency of Alpha have been proven.

Although Alpha can meet various need in DL, it's still in an experimental version and requires improvements. Presently, Alpha doesn't provide sequence models such as LSTM, GRU. The conv\deconv operators of cu32 are mainly designed for $32x$ channels and has best performance for $2^x$ channels, whose performance will degrade if not used accordingly.

The future work for Alpha could include:

a. More impls of EngineBase oriented at CPU, GPU, FGPA, TPU, etc.

b. Enrich Alpha's operators (including fused and novel operators).

c. Better memory-pool, data-structures and methods with lower cost.

d. Some experiments combing Alpha with Java big-data frameworks.